\documentclass{article}

\usepackage[sort&compress,numbers]{natbib}
\usepackage[preprint]{neurips_2026}

\usepackage[utf8]{inputenc} 
\usepackage[T1]{fontenc}    
\usepackage{hyperref}       
\usepackage{url}            
\usepackage{booktabs}       
\usepackage{amsfonts}       
\usepackage{nicefrac}       
\usepackage{microtype}      
\usepackage{xcolor}         

\usepackage{graphicx}
\usepackage{multirow}
\usepackage{amssymb}
\usepackage{amsmath}
\usepackage{subcaption}

\usepackage{xcolor}

\definecolor{gain}{RGB}{255,0,0}
\definecolor{drop}{RGB}{0,0,255}

\title{CoDeRNet: Selective Cross-Task Routing under Heterogeneous Supervision for Change Detection and Captioning}

\author{%
  Eunki Cho \quad Hyeon Bae Kim \quad Seong Tae Kim \\[2mm]
  Kyung Hee University, Republic of Korea \\
  \texttt{\{cekkec, hyeonbae.kim, st.kim\}@khu.ac.kr}
}

\begin{document}

\maketitle

\begin{abstract}
Change Detection \& Captioning (CDC) aims to jointly localize changed regions and describe their semantic content from bi-temporal remote sensing images. While recent approaches have demonstrated the effectiveness of unified modeling, they primarily rely on shared representations or jointly optimized architectures, leaving task-specific communication under heterogeneous supervision less explored. However, detection and captioning differ fundamentally in their supervision granularity, leading to heterogeneous representations with uneven reliability across tasks and spatial locations. In this paper, we reformulate CDC as a selective information routing problem between heterogeneous task branches, where the key challenge is to determine when and where information should be transferred. To this end, we propose CoDeRNet, a unified framework that preserves task-specific representations while enabling bidirectional exchange of complementary information. CoDeRNet introduces a \textbf{Co}nfidence-\textbf{De}composed \textbf{R}outing (CoDeR) mechanism, which decomposes the routing condition into receiver-side need for support and sender-side reliability cues to guide selective transfer. Experiments on LEVIR-MCI and WHU-CDC show that CoDeRNet achieves strong overall CDC performance, particularly improving caption generation while preserving competitive change localization. Extensive analyses further show that the benefits arise not merely from joint learning or generic cross-task communication, but from explicitly modeling selective cross-task information routing. These results highlight the importance of conditional information transfer for unified CDC under heterogeneous supervision.
\end{abstract}

\section{Introduction}
\label{sec:Introduction}

Understanding changes between bi-temporal remote sensing observations is a fundamental problem in remote sensing, with broad applications including urban planning~\cite{huang2013building}, natural disaster assessment~\cite{xu2019building}, and environmental monitoring~\cite{kennedy2009remote}. Two representative tasks have been widely studied in this context: Change Detection (CD)~\cite{daudt2018fully,fang2021snunet,bandara2022transformer,li2022remote,li2023lightweight,han2023hanet,chen2024changemamba}, which localizes changed regions at the pixel level, and Change Captioning (CC)~\cite{park2019robust,liu2022remote, liu2023progressive,liu2023decoupling,chang2023changes,sun2024lightweight,zhou2024single,sun2025mask}, which generates natural-language descriptions of semantic changes. Although their output formats differ, both tasks are grounded in the same underlying change event and provide complementary views of the same phenomenon. Motivated by this commonality, recent studies have introduced the unified setting of Change Detection \& Captioning (CDC)~\cite{shi2024multitask, liu2024change}. This unified setting is particularly appealing because it allows a single model to jointly reason about both where a change occurs and what the change means, offering a more complete form of change interpretation than either task alone.

However, these tasks differ fundamentally in their supervision granularity. Detection is densely supervised at the pixel level, whereas captioning is weakly supervised at the sentence level. 
This heterogeneity introduces a structural mismatch in how information is learned and represented across the two tasks. Existing CDC approaches primarily rely on shared representations or jointly optimized architectures~\cite{shi2024multitask, liu2024change}, establishing a unified framework for learning spatial localization and semantic description together. While this paradigm is effective, it mainly emphasizes joint modeling of the two tasks, leaving the role of task-specific communication under heterogeneous supervision less explored. Importantly, uniform interaction or jointly learned interaction mechanisms implicitly assume that cross-task information is equally reliable and equally useful across spatial locations and tasks. However, under heterogeneous supervision, task representations often exhibit asymmetric reliability and support requirements, making unconditional information transfer suboptimal.
Therefore, the central challenge is to determine how complementary information should be selectively transferred between heterogeneous task branches, rather than treating CDC merely as a joint optimization problem.

This perspective is particularly important in CDC because the usefulness of cross-task information can vary across both tasks and spatial locations. For example, detection-oriented features may provide strong spatial evidence around changed regions and boundaries, but they have limited capacity to explicitly organize such evidence into sentence-level semantic descriptions. Conversely, captioning-oriented features may capture global context and semantic relations, but sentence-level supervision does not explicitly constrain their spatial precision. This suggests that cross-task communication in CDC should be spatially and task-wise conditional, rather than uniformly applied across branches. Each branch should receive complementary information only when its own representation is insufficient and the other branch provides reliable supporting evidence.

Based on this motivation, we reformulate CDC as a selective information routing problem between heterogeneous task branches. To this end, we propose CoDeRNet, a unified framework that preserves task-specific branch identities while enabling bidirectional exchange of complementary information. CoDeRNet performs selective transfer via Confidence-Decomposed Routing (CoDeR), which decomposes the routing condition into receiver-side need for support and sender-side reliability cues. An overview of the proposed framework is shown in Fig.~\ref{fig:overview}.

We demonstrate that CoDeRNet achieves strong overall CDC performance on representative benchmarks, including LEVIR-MCI and WHU-CDC, with notable gains in caption generation while maintaining competitive change localization. Moreover, through systematic analyses of joint learning, routing directionality, mask decomposition, and alternative interaction operators, we show that the gains arise not merely from joint learning or generic interaction, but from explicitly modeling selective cross-task routing. The code is publicly available at \url{https://github.com/cekkec/CoDeRNet}.

\noindent The main contributions of this paper are summarized as follows.

\begin{itemize}
\item We reformulate CDC as a selective cross-task information routing problem under heterogeneous supervision, where complementary information should be transferred in a conditional and task-dependent manner.
\item We propose CoDeRNet, a unified CDC framework that preserves the distinct roles of detection and captioning while enabling task-aware bidirectional routing. Its core mechanism, Confidence-Decomposed Routing (CoDeR), guides selective transfer by decomposing the routing condition into receiver-side need for support and sender-side reliability cues.
\item We validate CoDeRNet on representative CDC benchmarks, including LEVIR-MCI and WHU-CDC, and demonstrate through systematic analyses of joint learning, routing directionality, mask decomposition, and alternative interaction operators the effectiveness of selective cross-task routing under heterogeneous supervision.
\end{itemize}

\section{Related Work}
\label{sec:Related Work}

\textbf{Remote Sensing Change Detection.}
Remote sensing change detection (RSCD) aims to identify meaningful changes from bi-temporal images of the same geographic area. Deep RSCD methods have evolved from CNN-based fully convolutional and Siamese encoder-decoder architectures~\cite{daudt2018fully, fang2021snunet, chen2021adversarial, zheng2021change, li2022remote, codegoni2023tinycd, li2023lightweight, han2023hanet} to Transformer-based models that capture long-range dependencies and global context~\cite{chen2021remote, bandara2022transformer, fang2023changer, noman2024remote, zhang2024bifa, zheng2024single, rolih2025change}. These studies have improved change localization by enhancing multi-scale representation learning, temporal feature comparison, and contextual reasoning over bi-temporal observations. More recently, Mamba-based architectures have been explored for efficient spatiotemporal modeling~\cite{chen2024changemamba, zhang2025cdmamba}, further expanding RSCD toward scalable and globally aware change representation.

\textbf{Remote Sensing Image Change Captioning.}
Remote sensing image change captioning (RSICC) aims to generate natural-language descriptions of semantic changes from bi-temporal images. Existing methods have improved change captioning through encoder-decoder architectures~\cite{park2019robust}, Transformer-based cross-temporal modeling and multi-scale change representation~\cite{liu2022remote, liu2023progressive}, pretrained language models and decoupled reasoning~\cite{liu2023decoupling}, attentive or lightweight architectures~\cite{chang2023changes, sun2024lightweight}, bi-temporal pretraining~\cite{zhou2024single}, and diffusion-based change modeling~\cite{sun2025mask}. These approaches have progressively strengthened the ability to localize visual change cues, model temporal differences, and translate them into semantically coherent descriptions.

\textbf{Remote Sensing Change Detection $\&$ Captioning.} Remote sensing change detection and captioning has recently emerged as a unified task for comprehensive change interpretation, combining pixel-level localization with semantic-level change description. MCINet~\cite{liu2024change} introduced LEVIR-MCI and proposed a dual-branch framework that jointly performs change detection and change captioning. CTMTNet~\cite{shi2024multitask} further formalized this direction as the RS-CDC task and developed a shared multitask architecture with task-specific decoders. These studies established unified CDC as a promising direction for jointly modeling spatial and semantic change information.

\section{Method}
\label{sec:Method}

\subsection{Overview}
\begin{figure}[t]
    \centering
    \includegraphics[width=\linewidth]{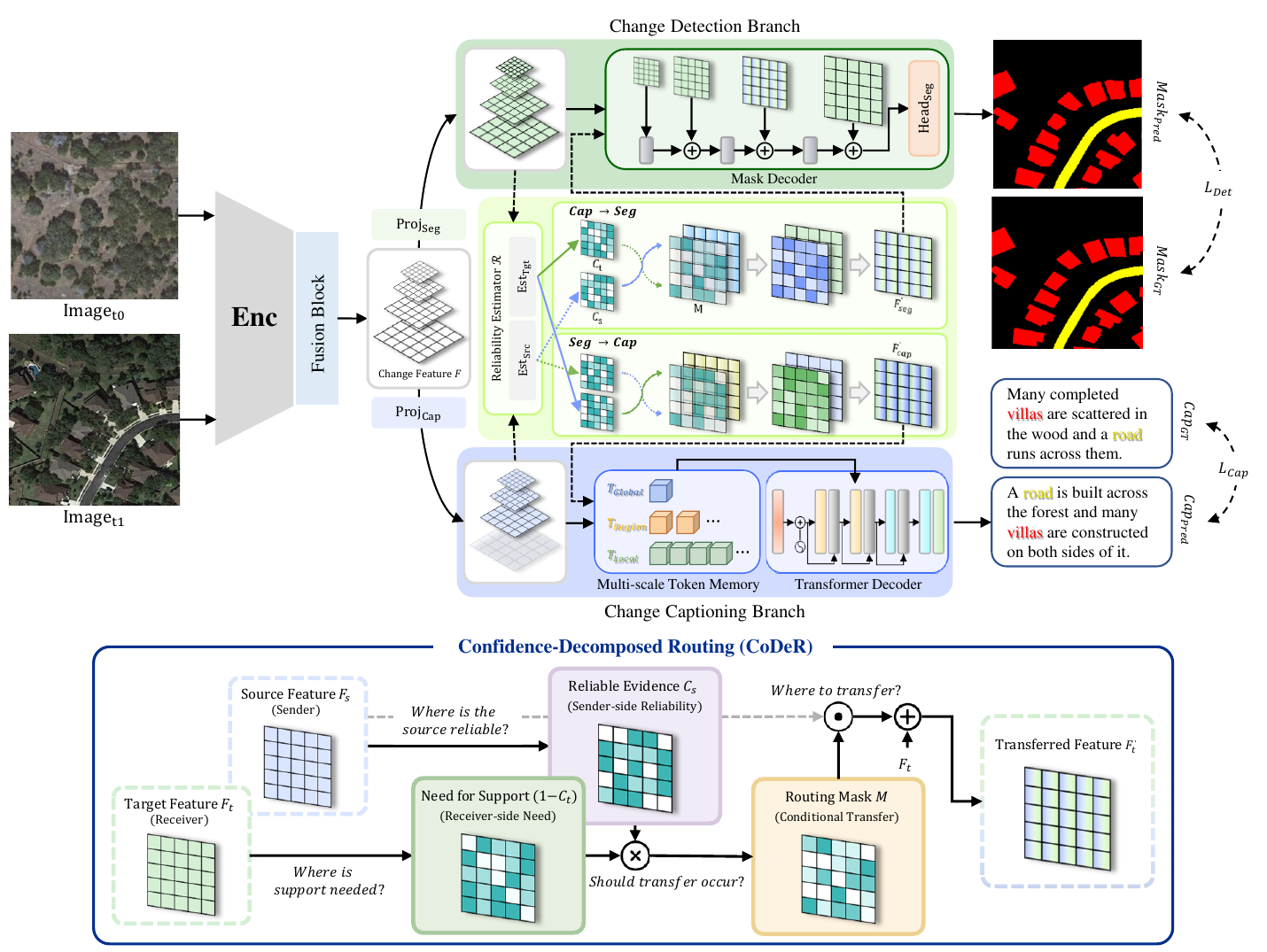}
\caption{Overview of CoDeRNet. Shared change features are transformed into task-specific detection and captioning features, followed by Confidence-Decomposed Routing (CoDeR) for selective bidirectional transfer before task-specific decoding.}
    \label{fig:overview}
\end{figure}

CoDeRNet addresses remote sensing change detection and change captioning within a unified yet selective information routing framework. Given a bi-temporal image pair $(I_{t_0}, I_{t_1})$, the model first extracts multi-scale visual features from the two images and constructs shared change representations. These shared change features are then passed through branch-specific interfaces that provide separate feature spaces for detection and captioning. On top of these branch-specific features, Confidence-Decomposed Routing (CoDeR) selectively routes complementary information between the two branches. Finally, task-specific decoders produce the change mask and the change description, respectively. As illustrated in Fig.~\ref{fig:overview}, the entire framework is optimized end-to-end under a joint training objective.

\subsection{Shared Change Representation}
Given a bi-temporal image pair $(I_{t_0}, I_{t_1})$, where $I_{t_0}$ and $I_{t_1}$ denote the earlier and later observations, respectively, we first extract multi-scale visual features from the two images using a shared encoder. Let $F_{t_0}^{(s)}, F_{t_1}^{(s)} \in \mathbb{R}^{B \times C_s \times H_s \times W_s}$ denote the features of the two observations at scale $s$. To construct a shared representation of temporal change, we combine the two temporal features with two complementary change cues: a magnitude change cue, which captures the strength of change, and a signed change cue, which preserves directional temporal differences. The resulting change representation at each scale is defined as
\[
F^{(s)} = \mathrm{Conv}_{1\times1}\big([F_{t_0}^{(s)}, F_{t_1}^{(s)}, F_{|\Delta|}^{(s)}, F_{\Delta}^{(s)}]\big),
\]
where $F_{|\Delta|}^{(s)} = |F_{t_1}^{(s)} - F_{t_0}^{(s)}|$ and $F_{\Delta}^{(s)} = F_{t_1}^{(s)} - F_{t_0}^{(s)}$ denote the magnitude and signed change cues.

\subsection{Branch-Specific Feature Interface}
Before cross-task routing, we instantiate branch-specific feature interfaces from the shared change representation. This step provides separate feature spaces in which CoDeR can estimate branch-dependent routing cues and apply residual information transfer:
\[
F_{\mathrm{seg}}^{(s)}=\phi_{\mathrm{seg}}^{(s)}(F^{(s)}), \qquad
F_{\mathrm{cap}}^{(s)}=\phi_{\mathrm{cap}}^{(s)}(F^{(s)}),
\]
where $\phi_{\mathrm{seg}}^{(s)}$ and $\phi_{\mathrm{cap}}^{(s)}$ are lightweight projection layers. These interfaces provide the feature spaces on which selective information routing is performed, while the actual cross-task transfer is handled by CoDeR. The same interfaces are used for all compared variants, so that subsequent comparisons focus on the routing mechanism rather than differences in branch construction.

\subsection{Confidence-Decomposed Routing}
After branch-specific interfaces are constructed, the two branches encode different yet complementary aspects of the same change. The detection branch tends to provide stronger cues for modeling where a change occurs, whereas the captioning branch tends to provide richer cues for reasoning about what the change means. Rather than directly fusing the two branches, we explicitly model where cross-task transfer is likely to be beneficial. Consider a target feature map $F_t \in \mathbb{R}^{B \times C \times H \times W}$ from the receiving branch and a source feature map $F_s \in \mathbb{R}^{B \times C \times H \times W}$ from the sending branch. CoDeR estimates target-side sufficiency and source-side reliability cues using lightweight routing estimators and derives a routing mask as
\[
C_t=\mathcal{R}_t(F_t), \qquad
C_s=\mathcal{R}_s(F_s), \qquad
M=(1-C_t)\cdot C_s,
\]
where $\mathcal{R}_t(\cdot)=\sigma(h_t(\cdot))$ and $\mathcal{R}_s(\cdot)=\sigma(h_s(\cdot))$ are lightweight routing estimators implemented by $1\times1$ convolution heads with sigmoid activation, and $C_t, C_s, M \in [0,1]^{B \times 1 \times H \times W}$. Here, $C_t$ and $C_s$ are not calibrated predictive probabilities, but task-dependent routing cues learned to estimate where cross-task transfer is expected to be complementary. The term $(1-C_t)$ represents the receiver-side need for support, while $C_s$ represents the sender-side reliability cue. Their multiplicative combination acts as a soft conjunction, making the routing mask large only when the receiving branch requires support and the sending branch provides reliable evidence. In this way, CoDeR routes information conditionally rather than applying cross-task transfer uniformly across all spatial locations.

The source feature is then mapped to the target feature space through a lightweight adapter $g$ and selectively injected via residual transfer:
\[
F_t' = F_t + M \odot g(F_s),
\]
where $\odot$ denotes element-wise multiplication with spatial broadcasting. When $M$ is small, the original target representation is largely preserved; when $M$ is large, the target branch receives complementary evidence from the source branch. This selective residual update allows CoDeRNet to preserve branch-specific representations while refining regions that may benefit from cross-task support.

CoDeR is applied bidirectionally, i.e., from captioning to detection and from detection to captioning, with independent routing estimators and adapters for the two directions:
\[
F_{\mathrm{seg}}^{\prime(s_g)}
=
\mathrm{CoDeR}_{\mathrm{cap}\rightarrow\mathrm{seg}}
\big(
F_{\mathrm{seg}}^{(s_g)},
F_{\mathrm{cap}}^{(s_g)}
\big),
\quad
F_{\mathrm{cap}}^{\prime(s_c)}
=
\mathrm{CoDeR}_{\mathrm{seg}\rightarrow\mathrm{cap}}
\big(
F_{\mathrm{cap}}^{(s_c)},
F_{\mathrm{seg}}^{(s_c)}
\big).
\]
Within each direction, the source and target features are taken from the same spatial scale, which allows the routing mask to be applied directly without additional resizing. The design is asymmetric across the two routing directions: the detection branch receives routed information at a relatively higher-resolution scale, whereas the captioning branch receives routed information at a lower-resolution scale. This design reflects the different spatial granularity requirements of the two tasks.

\begin{table*}[t]
\caption{Comparison on LEVIR-MCI~\cite{liu2024change}. ``Change Detection'' and ``Change Captioning'' denote single-task models trained for only one task, whereas ``Change Detection \& Captioning'' denotes unified models that perform both tasks jointly within a single framework. \textbf{Bold} and \underline{underlined} values indicate the best and second-best results, respectively.}
\label{tab:comparison_levir}
\centering
\small
\resizebox{\textwidth}{!}{%
\begin{tabular}{llcccccccc}
\toprule
\multirow{2}{*}{Task} & \multirow{2}{*}{Method} & \multicolumn{8}{c}{Metrics} \\
\cmidrule(lr){3-10}
& & mIoU & BLEU-1 & BLEU-2 & BLEU-3 & BLEU-4 & METEOR & ROUGE-L & CIDEr-D \\
\midrule

\multirow{6}{*}{\shortstack[c]{Change\\Detection}}
& SNUNet~\cite{fang2021snunet}
& 82.53 & - & - & - & - & - & - & - \\
& ChangeFormer~\cite{bandara2022transformer}
& 84.35 & - & - & - & - & - & - & - \\
& TFI-GR~\cite{li2022remote}
& 79.34 & - & - & - & - & - & - & - \\
& A2Net~\cite{li2023lightweight}
& 80.82 & - & - & - & - & - & - & - \\
& HANet~\cite{han2023hanet}
& 71.23 & - & - & - & - & - & - & - \\
& ChangeMamba~\cite{chen2024changemamba}
& \underline{87.61} & - & - & - & - & - & - & - \\
\midrule

\multirow{9}{*}{\shortstack[c]{Change\\Captioning}}
& DUDA~\cite{park2019robust}             & - & 81.44 & 72.22 & 64.24 & 57.79 & 37.15 & 71.04 & 124.32 \\
& MCCFormers~\cite{qiu2021describing}  & - & 82.16 & 72.95 & 65.42 & 59.41 & 38.26 & 72.10 & 128.34 \\
& RSICCformer~\cite{liu2022remote}       & - & 84.11 & 75.40 & 68.01 & 61.93 & 38.79 & 73.02 & 131.40 \\
& Prompt-CC~\cite{liu2023decoupling}     & - & 83.66 & 75.73 & 69.10 & 63.54 & 38.82 & 73.72 & 136.44 \\
& Chg2Cap~\cite{chang2023changes}        & - & 85.14 & 76.91 & 69.86 & 64.09 & 39.83 & 74.62 & 135.41 \\
& SparseFocus~\cite{sun2024lightweight} & - & 84.56 & 75.87 & 68.64 & 62.87 & 39.93 & 74.69 & 137.05 \\
& SEN~\cite{zhou2024single}              & - & 85.10 & 77.05 & 70.01 & 64.09 & 39.59 & 74.57 & 136.02 \\
& MaskApproxNet~\cite{sun2025mask}       & - & 85.90 & 77.12 & 70.72 & 64.32 & 39.91 & 75.67 & 137.71 \\

\midrule

\multirow{3}{*}{\shortstack[c]{Change\\Detection\\$\&$\\Captioning}}

& CTMTNet~\cite{shi2024multitask}
& 85.37 & 85.16 & 77.24 & 70.13 & 64.16 & 39.34 & 74.03 & 134.50 \\
& MCINet~\cite{liu2024change}
& 86.43 & \underline{86.68} & \underline{78.75} & \underline{71.74} & \underline{65.95} & \underline{40.80} & \underline{75.96} & \underline{140.29} \\

\cmidrule(lr){2-10}  
& CoDeRNet (Ours) & \textbf{87.79} & \textbf{87.25} & \textbf{79.42} & \textbf{72.26} & \textbf{66.27} & \textbf{41.08} & \textbf{76.36} & \textbf{141.79} \\

\bottomrule
\end{tabular}%
}
\end{table*}

\subsection{Task-Specific Decoders}
After branch-specific interfaces and CoDeR-based routing, each branch maintains a multi-scale feature hierarchy. CoDeR is applied only at a designated receiving scale for each branch, while the remaining scales are preserved as branch-specific features. The resulting hierarchies are then decoded by task-specific heads that reflect the distinct granularity and inductive bias required by change detection and change captioning. For notational simplicity, we use the primed feature notation to denote the branch features after CoDeR. At scales without CoDeR, the primed features are identical to the original branch features.

\paragraph{Segmentation Decoder.}
Change detection requires spatially precise prediction with progressive recovery of spatial detail. We therefore use an FPN-style top-down decoder~\cite{lin2017feature} to aggregate the multi-scale segmentation branch features. The decoder progressively combines coarse semantic features with higher-resolution features and produces pixel-wise class logits:
\[
\hat{M}
=
\mathcal{D}_{\mathrm{seg}}
\big(\{F_{\mathrm{seg}}^{\prime(s)}\}_{s \in \mathcal{S}}\big)
\in \mathbb{R}^{B \times K \times H \times W},
\]
where \(\mathcal{S}\) denotes the set of feature scales and \(K\) denotes the number of change classes.

\paragraph{Captioning Decoder.}
Unlike segmentation, caption generation requires organizing change evidence into semantic units. We therefore construct a multi-granularity memory from the caption-oriented branch features:
\[
\mathcal{M} = [\,t_{\mathrm{glob}};\, T_{\mathrm{reg}};\, T_{\mathrm{loc}}\,],
\qquad
\hat{p}_{t+1}
=
\mathcal{D}_{\mathrm{cap}}(y_{1:t}^{*}, \mathcal{M}).
\]
Here, \(t_{\mathrm{glob}}\), \(T_{\mathrm{reg}}\), and \(T_{\mathrm{loc}}\) denote global, region-level, and local tokens that capture overall context, intermediate semantic units, and fine spatial evidence, respectively. The Transformer decoder \(\mathcal{D}_{\mathrm{cap}}\) predicts the next-token distribution using the caption prefix and the multi-granularity memory.

\begin{table*}[t]
\caption{Comparison on WHU-CDC~\cite{shi2024multitask}. ``Change Detection'' and ``Change Captioning'' denote single-task models trained for only one task, whereas ``Change Detection \& Captioning'' denotes unified models that perform both tasks jointly within a single framework. \textbf{Bold} and \underline{underlined} values indicate the best and second-best results, respectively.}
\label{tab:comparison_whu}
\centering
\small
\resizebox{\textwidth}{!}{%
\begin{tabular}{llcccccccc}
\toprule
\multirow{2}{*}{Task} & \multirow{2}{*}{Method} & \multicolumn{8}{c}{Metrics} \\
\cmidrule(lr){3-10}
& & IoU & BLEU-1 & BLEU-2 & BLEU-3 & BLEU-4 & METEOR & ROUGE-L & CIDEr-D \\
\midrule

\multirow{6}{*}{\shortstack[c]{Change\\Detection}}
& SNUNet~\cite{fang2021snunet}
& 74.81 & - & - & - & - & - & - & - \\
& ChangeFormer~\cite{bandara2022transformer}
& 77.88 & - & - & - & - & - & - & - \\
& TFI-GR~\cite{li2022remote}
& 77.60 & - & - & - & - & - & - & - \\
& A2Net~\cite{li2023lightweight}
& 80.43 & - & - & - & - & - & - & - \\
& HANet~\cite{han2023hanet}
& 79.95 & - & - & - & - & - & - & - \\
& ChangeMamba~\cite{chen2024changemamba}
& \underline{89.69} & - & - & - & - & - & - & - \\

\midrule

\multirow{10}{*}{\shortstack[c]{Change\\Captioning}}
& DUDA~\cite{park2019robust}             & - & 79.04 & 69.53 & 61.57 & 55.64 & 34.29 & 68.98 & 121.85 \\
& MCCFormers~\cite{qiu2021describing}  & - & 81.12 & 75.04 & 69.95 & 65.34 & 42.11 & 78.52 & 147.09 \\
& RSICCformer~\cite{liu2022remote}       & - & 80.05 & 74.24 & 69.61 & 66.54 & 42.65 & 73.91 & 133.44 \\
& Prompt-CC~\cite{liu2023decoupling}     & - & 81.12 & 73.96 & 67.22 & 61.45 & 36.99 & 71.88 & 134.50 \\
& Chg2Cap~\cite{chang2023changes}      & - & 77.27 & 71.33 & 65.10 & 58.44 & 39.95 & 78.67 & \underline{160.64} \\
& SparseFocus~\cite{sun2024lightweight}  & - & 81.17 & 72.90 & 66.06 & 60.27 & 37.34 & 72.63 & 134.64 \\
& SEN~\cite{zhou2024single}              & - & 80.60 & 74.64 & 67.69 & 61.97 & 36.76 & 71.70 & 133.57 \\
& MaskApproxNet~\cite{sun2025mask}       & - & 81.34 & 75.68 & 71.16 & 67.73 & 43.89 & 75.41 & 135.31 \\

\midrule

\multirow{3}{*}{\shortstack[c]{Change\\Detection\\$\&$\\Captioning}}
& CTMTNet~\cite{shi2024multitask}
& 81.54 & 83.56 & 77.66 & 72.76 & 69.00 & 45.39 & 79.23 & 149.40 \\
& MCINet~\cite{liu2024change}
& 88.23 & \textbf{87.82} & \textbf{82.84} & \textbf{78.66} & \textbf{75.49} & \underline{49.03} & \underline{82.54} & 159.57 \\

\cmidrule(lr){2-10}
& CoDeRNet (Ours) & \textbf{90.25} & \underline{87.44} & \underline{82.38} & \underline{77.85} & \underline{74.17} & \textbf{49.04} & \textbf{83.58} & \textbf{163.37} \\

\bottomrule
\end{tabular}%
}
\end{table*}

\subsection{Training Objective}
CoDeRNet is trained in a multi-task setting for change detection and change captioning. Following prior CDC practice~\cite{liu2024change}, we use task-specific losses for the two branches and optimize them jointly:
\[
\mathcal{L}_{\mathrm{seg}}
=
\mathcal{L}_{\mathrm{CE}}(\hat{M}, M^{*})
+
\mathcal{L}_{\mathrm{Dice}}(\hat{M}, M^{*}),
\quad
\mathcal{L}_{\mathrm{cap}}
=
-
\sum_{t=1}^{T}
\log p\!\left(
y_t^{*}\mid y_{<t}^{*}, \mathcal{M}
\right).
\]
Here, $M^{*}$ denotes the ground-truth change mask. The segmentation branch is supervised by pixel-wise cross-entropy and Dice loss, while the captioning branch is trained with token-level cross-entropy under teacher forcing over valid non-padded positions.

For stable joint optimization, we normalize each loss by its detached current magnitude:
\[
\mathcal{L}_{\mathrm{total}}
=
\frac{\mathcal{L}_{\mathrm{seg}}}{\operatorname{sg}(\mathcal{L}_{\mathrm{seg}})+\epsilon}
+
\frac{\mathcal{L}_{\mathrm{cap}}}{\operatorname{sg}(\mathcal{L}_{\mathrm{cap}})+\epsilon},
\]
where $\operatorname{sg}(\cdot)$ denotes the stop-gradient operator and $\epsilon$ is a small constant for numerical stability. This normalization is used only as a training stabilization strategy to balance the scales of the two task losses.

\begin{figure}[t]
    \centering
    \includegraphics[width=\linewidth]{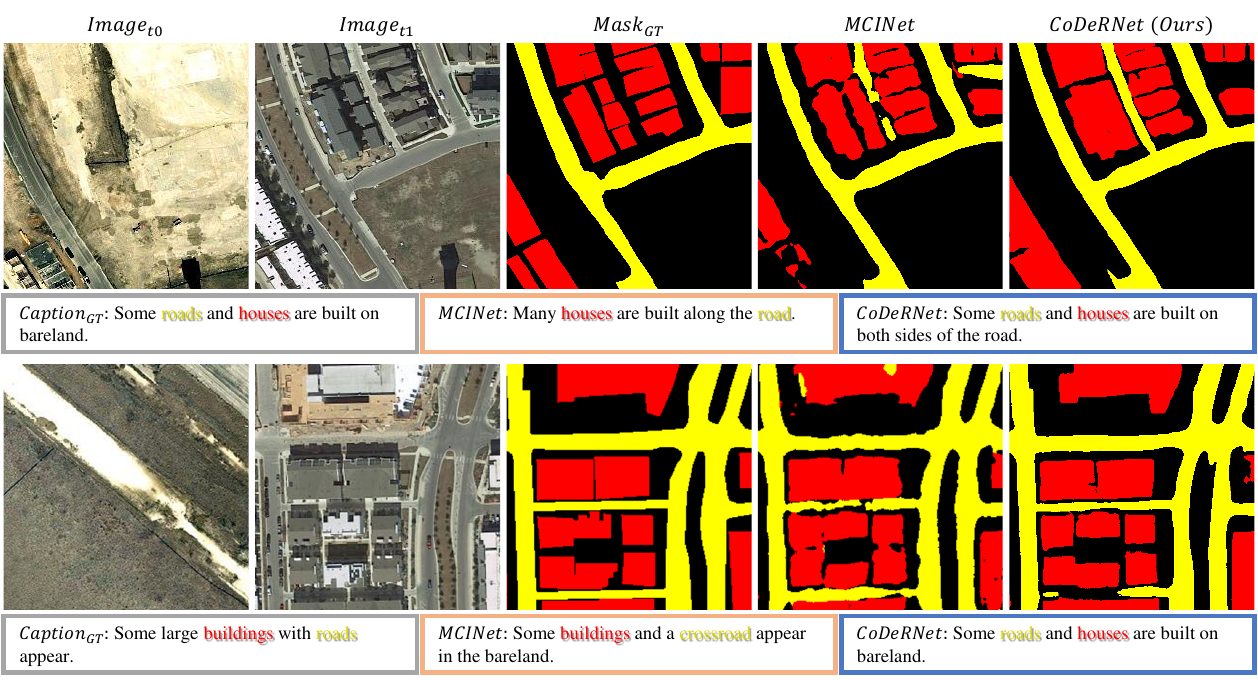}
\caption{Qualitative comparison on LEVIR-MCI. In the captions, building-related words (e.g., \textit{house}, \textit{building}) and road-related words (e.g., \textit{road}) are highlighted using colors consistent with the corresponding mask classes.}
    \label{fig:qualitative_levir}
\end{figure}

\section{Experiments}
\label{sec:Experiments}

\subsection{Datasets and Evaluation Metrics}
\textbf{Datasets.}
We evaluate CoDeRNet on two representative CDC benchmarks, LEVIR-MCI~\cite{liu2024change} and WHU-CDC~\cite{shi2024multitask}. LEVIR-MCI, extended from LEVIR-CC~\cite{liu2022remote}, contains 10,077 bi-temporal image pairs of size $256\times256$ at 0.5\,m resolution, each annotated with a multi-class change mask for buildings and roads and five human-written captions. Following the standard split, it is divided into 6,815 training, 1,333 validation, and 1,929 test samples.

WHU-CDC, derived mainly from WHU-CD~\cite{ji2018fully}, contains 7,434 bi-temporal image pairs of size $256\times256$ at 0.075\,m resolution, focusing on post-earthquake urban reconstruction in Christchurch, New Zealand. Each pair has a binary building-change mask and five captions describing multiple change categories, including buildings, parking lots, roads, vegetation, and water. The training, validation, and test sets contain 5,947, 743, and 744 samples, respectively. These datasets allow us to evaluate CoDeRNet under different levels of mask granularity and caption semantics.

Both datasets can contain multiple change events per image pair and cover multiple change categories in their annotations, although WHU-CDC masks are restricted to building changes.

\textbf{Evaluation Metrics.}
For change detection, we follow prior CDC benchmarks and report mIoU on LEVIR-MCI~\cite{liu2024change}, which provides multi-class change masks for buildings, roads, and no-change regions. On WHU-CDC~\cite{shi2024multitask}, which provides binary building-change masks, we report the IoU of the change class. For change captioning, we adopt standard language generation metrics, including BLEU-N (N = 1,2,3,4)~\cite{papineni2002bleu}, METEOR~\cite{banerjee2005meteor}, ROUGE-L~\cite{lin2004rouge}, and CIDEr-D~\cite{vedantam2015cider}, covering n-gram overlap, semantic matching, sequence similarity, and consensus-based caption quality.

\subsection{Implementation Details}

All experiments are implemented in PyTorch and conducted on a single NVIDIA RTX 4090 GPU. We use a ConvNeXt-Base backbone~\cite{liu2022convnet} with ImageNet-1K pretrained weights and apply it to the two temporal images in a weight-shared manner. The maximum caption length is set to 41 for LEVIR-MCI~\cite{liu2024change} and 26 for WHU-CDC~\cite{shi2024multitask}, and greedy autoregressive decoding is used during inference. For training, we use Adam~\cite{kingma2014adam} with a learning rate of $1\times10^{-4}$ and an effective batch size of 64, achieved through a micro-batch size of 16 and four gradient accumulation steps. The learning rate is decayed by a factor of 0.5 every 50 epochs, and the model is trained for 100 epochs on LEVIR-MCI and 150 epochs on WHU-CDC. We apply random horizontal and vertical flips together with $90^\circ$ rotations during training. When geometric augmentation is applied, direction-sensitive caption words are remapped accordingly. Detailed remapping rules are provided in Sec.~\ref{sec:add_ablation}. For both LEVIR-MCI and WHU-CDC, the input size is fixed to $256\times256$. For this setup, a single training run takes approximately 12 hours.

\subsection{Comparison with State-of-the-Art Methods}
We compare CoDeRNet with recent state-of-the-art methods on LEVIR-MCI~\cite{liu2024change} and WHU-CDC~\cite{shi2024multitask}, including single-task change detection models, single-task change captioning models, and unified change detection-captioning models. This comparison evaluates whether a unified framework can improve spatial localization and semantic description of changes at the same time. As shown in Table~\ref{tab:comparison_levir}, CoDeRNet achieves the best overall performance on LEVIR-MCI across both detection and captioning metrics. For example, CoDeRNet achieves BLEU-4 66.27, CIDEr-D 141.79, and mIoU 87.79 on LEVIR-MCI, showing stronger overall performance than the strongest prior unified model.

Table~\ref{tab:comparison_whu} shows a similar trend on WHU-CDC. CoDeRNet achieves the best IoU of 90.25 and improves consensus-oriented captioning metrics, obtaining the best CIDEr-D of 163.37 and ROUGE-L of 83.58, while remaining competitive on BLEU-based metrics.

Compared with prior unified CDC frameworks, CoDeRNet preserves strong localization accuracy while improving semantic description quality across both datasets, leading to a better overall balance between the two tasks. Overall, these results show that CoDeRNet effectively supports both spatial and semantic change understanding within a unified framework, while remaining robust across datasets with different change characteristics and annotation settings.

Qualitative comparisons in Fig.~\ref{fig:qualitative_levir} further show that the visual improvements are consistent with the quantitative results on LEVIR-MCI. Additional qualitative examples, including further results on LEVIR-MCI (Fig.~\ref{fig:qualitative_levir2}) and WHU-CDC (Fig.~\ref{fig:qualitative_whu}), are provided in Appendix.

\begin{table*}[t]
\caption{Ablation studies on LEVIR-MCI~\cite{liu2024change}. (a) Effect of joint training and explicit interaction, comparing single-task training, joint training without CoDeR, and joint training with CoDeR. (b) Effect of CoDeR directionality, comparing two single-direction settings with the bidirectional design. \textbf{Bold} indicates the best result in each column.}
\label{tab:ablation_direction}
\centering
\small

\begin{minipage}[t]{0.49\textwidth}
\centering
\textbf{(a) Joint training and CoDeR}
\vspace{2pt}

\resizebox{\textwidth}{!}{%
\begin{tabular}{cccccccc}
\toprule
Seg & Cap & CoDeR & mIoU & BLEU-4 & METEOR & ROUGE-L & CIDEr-D \\
\midrule
\checkmark & -          & -          & 87.03          & -              & -              & -              & -              \\
-          & \checkmark & -          & -              & 63.18          & 40.21          & 75.27          & 137.41         \\
\checkmark & \checkmark & -          & 87.67          & 64.41          & 39.93          & 74.98          & 136.18         \\
\checkmark & \checkmark & \checkmark & \textbf{87.79} & \textbf{66.27} & \textbf{41.08} & \textbf{76.36} & \textbf{141.79} \\
\bottomrule
\end{tabular}%
}
\end{minipage}
\hfill
\begin{minipage}[t]{0.49\textwidth}
\centering
\textbf{(b) CoDeR directionality}
\vspace{2pt}

\resizebox{\textwidth}{!}{%
\begin{tabular}{lccccc}
\toprule
Direction & mIoU & BLEU-4 & METEOR & ROUGE-L & CIDEr-D \\
\midrule
cap $\rightarrow$ seg     & 87.53          & 64.79          & 40.27          & 75.25          & 138.62 \\
seg $\rightarrow$ cap     & 87.60          & 64.96          & 40.95          & 76.33          & \textbf{141.95} \\
seg $\leftrightarrow$ cap & \textbf{87.79} & \textbf{66.27} & \textbf{41.08} & \textbf{76.36} & 141.79 \\
\bottomrule
\end{tabular}%
}
\end{minipage}

\end{table*}


\subsection{Ablation Study}

\noindent\textbf{Effect of Joint Training and CoDeR.}
We first examine whether selective routing provides additional benefits beyond simple joint optimization, as shown in Table~\ref{tab:ablation_direction}(a). Compared with single-task training and joint training without routing, the joint model with CoDeR achieves the best overall performance across both detection and captioning metrics. This result suggests that joint learning alone is helpful but not sufficient, and that explicit cross-task routing further improves unified change understanding.

\noindent\textbf{Effect of Routing Directionality.}
To examine how CoDeR should be applied between the two branches, we compare two single-direction settings with the bidirectional design. As shown in Table~\ref{tab:ablation_direction}(b), the bidirectional design achieves the best overall balance across metrics among the three compared settings. This suggests that, once cross-task routing is introduced, mutual information exchange between the two branches is more effective than one-way transfer alone.

\noindent\textbf{Effect of Routing Mask Decomposition.}
To validate the core design of CoDeR, we ablate the decomposed routing mask and compare it with simplified variants. As shown in Table~\ref{tab:CoDeR_effective}(a), the full formulation consistently performs best, while removing either the target-dependent or source-dependent term degrades performance. Replacing the decomposition with a single gate also results in weaker performance. These results show that the benefit of CoDeR does not arise merely from gating itself, but from explicitly decomposing the routing condition into receiver-side need for support and sender-side reliability cues.

\noindent\textbf{Effect of Interaction Operators.}
To verify that the gains of CoDeRNet come from the proposed routing design rather than simply adding cross-task interaction, we compare CoDeR with representative interaction operators in Table~\ref{tab:CoDeR_effective}(b), including co-attention~\cite{lu2016hierarchical,nguyen2018improved}, cross-attention~\cite{wei2020multi,yu2020deformable}, channel exchange~\cite{wang2020deep}, and gated modulation. Since these methods were originally designed for different tasks and architectures, we implement their core interaction principles within the same CoDeRNet pipeline while keeping all other components fixed. Detailed formulations are provided in Appendix~\ref{sec:appendix_interaction_formulation}. As shown in the table, generic interaction operators can improve cross-branch communication, but they do not explicitly determine whether the receiving branch needs support or whether the sending branch provides reliable evidence. In contrast, CoDeR structures cross-task communication as conditional routing under heterogeneous supervision, leading to stronger captioning performance while preserving competitive detection accuracy. These results show that the effectiveness of cross-task communication in CDC depends not only on exchanging information, but also on how the transfer condition is defined.

Further analyses on CoDeR and other design choices are provided in Appendix Secs.~\ref{sec:scale_selection}--\ref{sec:add_ablation}, where we examine the effect of routing scale, backbone, and additional architectural and training factors.

\begin{table*}[t]
\caption{Ablation studies on LEVIR-MCI~\cite{liu2024change} for analyzing the effectiveness of CoDeR. 
(a) Routing mask decomposition ablation compares CoDeR with reduced variants and a single-gate design.
(b) Interaction operator ablation compares CoDeR with representative cross-task communication operators under the same framework.  \textbf{Bold} and \underline{underline} indicate the best and second-best results in each column, respectively.}
\label{tab:CoDeR_effective}
\centering
\small

\begin{minipage}[t]{0.49\textwidth}
\centering
\textbf{(a) CoDeR mask decomposition}
\vspace{2pt}

\resizebox{\textwidth}{!}{%
\begin{tabular}{lccccc}
\toprule
Mask & mIoU & BLEU-4 & METEOR & ROUGE-L & CIDEr-D \\
\midrule
Sender reliability $(C_s)$              & 87.60 & 65.12 & 40.49 & 75.73 & 139.57 \\
Receiver need $(1-C_t)$           & 87.65 & 64.65 & 40.81 & 75.94 & 141.02 \\
Single gate                            & 87.59 & 65.46 & 40.27 & 75.54 & 138.19 \\
\midrule
Decomposed routing $((1-C_t)\cdot C_s)$   & \textbf{87.79} & \textbf{66.27} & \textbf{41.08} & \textbf{76.36} & \textbf{141.79} \\
\bottomrule
\end{tabular}%
}
\end{minipage}
\hfill
\begin{minipage}[t]{0.5\textwidth}
\centering
\textbf{(b) Interaction operators}
\vspace{2pt}

\resizebox{\textwidth}{!}{%
\begin{tabular}{lccccccc}
\toprule
Interaction & mIoU & BLEU-4 & METEOR & ROUGE-L & CIDEr-D & Params (M) & FLOPs (G) \\
\midrule
Co-Attn    & \textbf{87.84} & 64.15 & 40.64 & 75.44 & 139.33 & 102.8 & 67.46 \\
Cross-Attn & 87.70 & 65.01 & 40.60 & 75.53 & 138.87 & 103.2 & 67.71 \\
Exchange   & 87.53 & 65.15 & 40.61 & 75.47 & 139.12 & 102.9 & 67.46 \\
Gating     & 87.75 & \underline{65.66} & \underline{40.73} & \underline{76.09} & \underline{140.86} & 103.1 & 67.46 \\
CoDeR (Ours) & \underline{87.79} & \textbf{66.27} & \textbf{41.08} & \textbf{76.36} & \textbf{141.79} & 102.8 & 67.46 \\
\bottomrule
\end{tabular}%
}
\end{minipage}

\end{table*}

\subsection{Human Evaluation of Caption Grounding and Cross-Output Consistency}
\label{sec:human_evaluation}

Standard captioning metrics measure agreement with reference captions but do not directly establish whether generated descriptions faithfully reflect observable changes or agree with the model's spatial predictions. We therefore conduct a blind human evaluation using two complementary criteria: factual grounding, which assesses the correctness of a caption with respect to the annotated changes, and cross-output consistency, which assesses agreement between a caption and the change mask predicted by the same model.

We evaluate MCINet, our baseline without CoDeR, and CoDeRNet on 660 change-containing image pairs from the LEVIR-MCI test set. The samples are stratified by change composition (building only, road only, or both) and change area. Seven researchers participated in the blind evaluation, with model identities concealed. Each evaluator assesses all three systems on the same assigned set of 120 image pairs in two separate rounds. In the factual-grounding round, captions are evaluated against the ground-truth change masks, without access to predicted masks. In the cross-output-consistency round, captions are evaluated against the corresponding predicted change masks, while ground-truth masks are withheld. Both criteria are rated on a five-point scale, with higher scores indicating closer agreement between the caption and the mask provided in the corresponding evaluation round. Both criteria are restricted to building- and road-related changes covered by the annotations; mentions of categories outside this scope are treated as neutral. Thirty image pairs are shared among all seven evaluators to assess inter-rater agreement. Each round yields 840 ratings per system and 2,520 ratings across the three systems, resulting in 5,040 criterion scores across both rounds.

Table~\ref{tab:human_evaluation} reports the mean and standard deviation of the ratings. CoDeRNet achieves the highest mean scores on both criteria, improving factual grounding and cross-output consistency over MCINet by 0.138 and 0.181, respectively. Both improvements are statistically significant under the sign test, with $p=0.045$ for factual grounding and $p=0.0004$ for cross-output consistency. CoDeRNet also achieves slightly higher mean scores than the baseline without CoDeR on both criteria. On the shared evaluation subset, factual-grounding ratings yield Krippendorff's $\alpha=0.739$, while the reliability of the average ratings across the seven evaluators, measured by $\mathrm{ICC}(A,7)$, reaches 0.974 for factual grounding and 0.892 for cross-output consistency. Together, these results provide evidence beyond reference-based captioning metrics that CoDeRNet improves factual grounding and cross-output consistency over MCINet.



\begin{table}[t]
\centering
\caption{Blind human evaluation on the change-containing subset of LEVIR-MCI. Values are mean $\pm$ standard deviation on a five-point scale, with 840 ratings per system for each criterion across 660 unique image pairs. Higher scores indicate better factual grounding or cross-output consistency. \textbf{Bold} indicates the highest mean in each column.}
\label{tab:human_evaluation}
\small
\begin{tabular}{lcc}
\toprule
Model & Factual grounding & Cross-output consistency \\
\midrule
MCINet~\cite{liu2024change} & $3.520 \pm 1.38$ & $3.632 \pm 1.25$ \\
Ours (w/o CoDeR) & $3.645 \pm 1.22$ & $3.782 \pm 1.09$ \\
Ours (w/ CoDeR) & $\mathbf{3.658} \pm 1.27$ & $\mathbf{3.813} \pm 1.13$ \\
\bottomrule
\end{tabular}
\end{table}

\section{Discussion}
\label{sec:Discussion}

Our findings suggest that CDC highlights a broader challenge in multi-task learning under heterogeneous supervision, where dense prediction and language generation require different forms of representation. Rather than assuming uniform feature sharing, CoDeRNet shows that conditional information routing can provide a more suitable way to coordinate such heterogeneous task branches.

\noindent\textbf{Limitation.}
Although CoDeRNet improves both tasks within a unified CDC framework, the performance gains are asymmetric. While captioning benefits significantly from detection cues, the reverse impact is more constrained by the semantic gap. This suggests that spatially grounded cues from the detection branch are particularly effective for improving semantic change description, whereas the reverse direction provides a more moderate benefit. In addition, our evaluation is restricted to two bi-temporal remote-sensing CDC benchmarks, and the effectiveness of CoDeRNet in multitemporal scenarios and broader dense-prediction-and-language settings remains to be verified.
We also do not explicitly supervise the routing cues with region-language correspondence, which may limit semantic faithfulness in cases where visual changes are spatially subtle or linguistically ambiguous.

\noindent\textbf{Broader Impact.}
CoDeRNet may support applications such as urban monitoring, disaster assessment, and environmental analysis. However, incorrect change localization or captioning could mislead downstream decisions, so human verification is recommended in high-stakes applications.

\noindent\textbf{Future Work.}
A promising direction for future work is to strengthen semantic correspondence between the two branches more explicitly. For example, region-language alignment or class-aware semantic transfer may help cross-task routing reflect not only spatial relevance but also the semantic structure shared by the two tasks. Such designs may provide a more effective way to translate semantic cues into region-level predictions within CDC.

\section{Conclusion}
\label{sec:Conclusion}
In this paper, we revisit change detection and captioning not simply as a conventional joint learning problem, but as a conditional information routing problem under heterogeneous supervision. We proposed CoDeRNet, a unified framework that preserves task-specific representations while enabling selective bidirectional exchange of complementary information. At its core, Confidence-Decomposed Routing decomposes the routing condition into receiver-side need for support and sender-side reliability cues, allowing cross-task information to be transferred only when it is expected to be beneficial.
Experiments on LEVIR-MCI and WHU-CDC show that CoDeRNet improves semantic description quality while preserving strong localization accuracy. Through systematic analyses, we further show that the gains arise not merely from generic feature interaction, but from explicitly modeling selective cross-task communication. The key contribution of CoDeR is therefore not the introduction of another interaction operator, but the reformulation of CDC communication as a conditional routing problem under heterogeneous supervision. This emphasizes the need to model task-specific reliability and support requirements when coordinating branches with different supervision signals. We believe this perspective extends beyond remote sensing CDC to broader multi-task and multimodal learning settings involving heterogeneous supervision and task coordination. 

\section*{Acknowledgments}
This work was supported in part by InSpace Company Ltd. (Inspace-Kyung Hee Artificial Intelligence Research Institute), by the National Research Foundation of Korea (NRF) grant funded by the Korea government (MSIT) (No. RS-2024-00334321), by the MSIT (Ministry of Science and ICT), Korea, under the ITRC (Information Technology Research Center) support program (No. IITP-2026-RS-2023-00258649) supervised by the IITP (Institute for Information and Communications Technology Planning and Evaluation), and by Global - Learning \& Academic research institution for Master’s·PhD students, and Postdocs (G-LAMP) Program of the National Research Foundation of Korea (NRF) grant funded by the Ministry of Education(No. RS-2025-25442355). The authors declare no competing interests.

\newpage

{\small
\bibliographystyle{plainnat}
\bibliography{main}
}

\newpage

\section{Appendix}
\label{sec:Appendix}

\subsection{Scale Selection for CoDeR}
\label{sec:scale_selection}

We further examine how the receiving scale affects CoDeR. We first vary the routing scale in each direction while fixing the opposite direction to its default setting, as shown in Table~\ref{tab:scale}. We then compare the default asymmetric single-scale design with multi-scale CoDeR variants in Table~\ref{tab:multiscale_region}(a) to test whether broader cross-scale routing provides additional benefit.

Table~\ref{tab:scale}(a) shows that, for \emph{cap $\rightarrow$ seg}, applying CoDeR at $1/8$ yields the strongest overall performance, suggesting that the detection branch benefits most when caption-derived information is injected at a relatively higher-resolution stage. Table~\ref{tab:scale}(b) shows that, for \emph{seg $\rightarrow$ cap}, the best overall performance is obtained at $1/16$, indicating that the captioning branch benefits more from interaction at a relatively lower-resolution stage aligned with semantic abstraction.

Table~\ref{tab:multiscale_region}(a) further shows that the default asymmetric single-scale design remains the most consistent configuration overall. The fully multi-scale variant degrades all caption metrics, while the direction-specific multi-scale variants do not establish a clear advantage over the default. These results indicate that the effectiveness of CoDeR comes not from broader cross-scale transfer, but from selecting an appropriate routing scale for each direction.

Overall, the results support the asymmetric scale design of CoDeRNet: \emph{cap $\rightarrow$ seg} is most effective at a higher-resolution scale, whereas \emph{seg $\rightarrow$ cap} is most effective at a lower-resolution scale. More importantly, CoDeR should be applied selectively at an appropriate granularity rather than uniformly across all scales.

\subsection{Backbone-Level Generalizability of CoDeR}
\label{sec:backbone_and_CoDeR}
We further examine whether the effect of CoDeR is consistent across different encoder choices, rather than being tied to a specific backbone architecture. To this end, Table~\ref{tab:backbone_robustness} compares the baseline without CoDeR and CoDeRNet under the same backbone setting, across diverse backbone families with image-only pretraining (e.g., ConvNeXt~\cite{liu2022convnet}) or vision-language pretraining (e.g., TIPSv2~\cite{cao2026tipsv2}), as well as different model sizes within the ConvNeXt series.


As shown in Table~\ref{tab:backbone_robustness}(a), enabling CoDeR generally improves captioning performance across all tested backbone families, with consistent gains in BLEU-4, METEOR, ROUGE-L, and CIDEr-D in most cases. Although the effect on mIoU is smaller and not always positive, the overall trend indicates that the proposed routing mechanism is not limited to a particular encoder design. Table~\ref{tab:backbone_robustness}(b) shows a similar pattern across different ConvNeXt scales. From Tiny to Large, CoDeR consistently improves most captioning metrics, while detection performance is generally maintained and occasionally shows modest improvements. These results suggest that the benefit of CoDeR is broadly preserved across backbone families and model scales, supporting the general applicability of the proposed routing design.

\subsection{Additional Ablation Studies}
\label{sec:add_ablation}

\noindent\textbf{Effect of region-token pooling size.}
Table~\ref{tab:multiscale_region}(b) studies the pooling size used to construct region-level tokens in the caption branch. Compared with using no region-level tokens, introducing pooled region tokens generally improves both detection and captioning performance, confirming the usefulness of intermediate-granularity representations. Among the tested settings, a pooling size of 2 gives the best overall performance on most metrics. Although a pooling size of 4 remains competitive, it does not surpass the overall performance of size 2, suggesting that an appropriate region granularity is more effective than overly coarse pooling.

\noindent\textbf{Effect of augmentation type.}
Table~\ref{tab:aug_comp}(a) analyzes the effect of geometric augmentation and directional caption remapping. Applying geometric augmentation alone improves several captioning metrics, but the gains are not consistent across all metrics. In contrast, combining geometric augmentation with caption remapping yields the best overall results for both detection and captioning. This indicates that, when direction-sensitive transformations are applied to the input images, maintaining caption-image consistency through remapping (see Table~\ref{tab:caption_remap_all}) is important for effective training.

\noindent\textbf{Effect of change fusion composition.}
Table~\ref{tab:aug_comp}(b) examines how the composition of the shared change representation affects performance. Starting from the temporal features $\{t_0,t_1\}$, adding the signed change cue $\Delta$ already improves the captioning-related metrics, indicating that directional temporal difference provides useful semantic information. Further adding the magnitude cue $|\Delta|$ leads to the best overall performance across all reported metrics. These results suggest that the signed and magnitude change cues play complementary roles, and that combining both is more effective than using temporal features alone.

\begin{table*}[t]
\caption{Ablation of the CoDeR routing scale on LEVIR-MCI~\cite{liu2024change}. Each subtable reports performance when same-scale routing is applied at different receiving scales in one direction, while the opposite direction is fixed to its default setting. \textbf{Bold} indicates the best result in each column.}
\label{tab:scale}
\centering
\small
\begin{minipage}{0.49\textwidth}
\centering
\subcaption{\textbf{Cap $\rightarrow$ seg: routing scale}}
\resizebox{\textwidth}{!}{%
\begin{tabular}{lccccc}
\toprule
Scale & mIoU & BLEU-4 & METEOR & ROUGE-L & CIDEr-D \\
\midrule
1/4              & 87.75          & 64.93          & 40.62          & 75.61          & 139.84 \\
$1/8$ & \textbf{87.79} & \textbf{66.27} & \textbf{41.08} & \textbf{76.36} & \textbf{141.79} \\
1/16             & 87.48          & \textbf{66.27} & 40.49          & 76.17          & 139.86 \\
1/32             & 87.78          & 65.72          & 40.71          & 75.93          & 139.80 \\
\bottomrule
\end{tabular}%
}
\end{minipage}
\hfill
\begin{minipage}{0.49\textwidth}
\centering
\subcaption{\textbf{Seg $\rightarrow$ cap: routing scale}}
\resizebox{\textwidth}{!}{%
\begin{tabular}{lccccc}
\toprule
Scale & mIoU & BLEU-4 & METEOR & ROUGE-L & CIDEr-D \\
\midrule
1/8              & 87.76          & 65.54          & 40.38          & 75.22          & 138.04 \\
$1/16$ & \textbf{87.79} & \textbf{66.27} & \textbf{41.08} & \textbf{76.36} & \textbf{141.79} \\
1/32             & 87.59          & 66.13          & 40.23          & 75.50          & 138.45 \\
\bottomrule
\end{tabular}%
}
\end{minipage}
\end{table*}

\begin{table*}[t]
\caption{Robustness of CoDeRNet across backbone choices on LEVIR-MCI~\cite{liu2024change}. (a) Backbone-family ablation: performance changes with and without CoDeR across different backbones. (b) Backbone-size ablation: performance changes with and without CoDeR across ConvNeXt variants from Tiny to Large. \textbf{Bold} indicates the best result in each column. In the CoDeRNet rows, colored values indicate improvements (\textcolor{gain}{gain}) or degradations (\textcolor{drop}{drop}) relative to the corresponding baseline. B4, MET., R-L, and CID. denote BLEU-4, METEOR, ROUGE-L, and CIDEr-D, respectively.}
\label{tab:backbone_robustness}
\centering
\small

\begin{minipage}[t]{0.49\textwidth}
\centering
\textbf{(a) Backbone family ablation}
\vspace{2pt}

\resizebox{\textwidth}{!}{%
\begin{tabular}{llccccc}
\toprule
Backbone & CoDeR & mIoU & BLEU-4 & MET. & R-L & CID. \\
\midrule
\multirow{2}{*}{ConvNeXt-B~\cite{liu2022convnet}}
 & $\times$     & 87.67 & 64.41 & 39.93 & 74.98 & 136.18 \\
 & $\checkmark$ & \textbf{\textcolor{gain}{87.79}} & \textbf{\textcolor{gain}{66.27}} & \textbf{\textcolor{gain}{41.08}} & \textbf{\textcolor{gain}{76.36}} & \textbf{\textcolor{gain}{141.79}} \\
\cmidrule(lr){2-7}
\multirow{2}{*}{SwinV2-B~\cite{liu2022swin}}
 & $\times$     & 87.49 & 63.55 & 40.24 & 75.26 & 136.93 \\
 & $\checkmark$ & \textcolor{drop}{87.34} & \textcolor{gain}{64.45} & \textcolor{gain}{40.30} & \textcolor{gain}{75.31} & \textcolor{gain}{138.20} \\
\cmidrule(lr){2-7}
\multirow{2}{*}{MiT-B5~\cite{xie2021segformer}}
 & $\times$     & 87.74 & 63.92 & 40.15 & 75.49 & 137.54 \\
 & $\checkmark$ & \textcolor{gain}{87.78} & \textcolor{gain}{65.38} & \textcolor{gain}{40.54} & \textcolor{gain}{75.54} & \textcolor{gain}{140.28} \\
\cmidrule(lr){2-7}
\multirow{2}{*}{DaViT-B~\cite{ding2022davit}}
 & $\times$     & 87.27 & 63.32 & 39.70 & 74.94 & 136.85 \\
 & $\checkmark$ & \textcolor{gain}{87.37} & \textcolor{gain}{64.59} & \textcolor{gain}{40.34} & \textcolor{gain}{75.24} & \textcolor{gain}{137.03} \\
\cmidrule(lr){2-7}
\multirow{2}{*}{EfficientNetV2-M~\cite{tan2021efficientnetv2}}
 & $\times$     & 87.05 & 63.31 & 40.31 & 75.66 & 137.97 \\
 & $\checkmark$ & \textcolor{gain}{87.10} & \textcolor{gain}{65.17} & \textcolor{gain}{40.72} & \textcolor{gain}{75.99} & \textcolor{gain}{140.12} \\
 \cmidrule(lr){2-7}
\multirow{2}{*}{TIPSv2-B/14~\cite{cao2026tipsv2}}
 & $\times$     & 86.56 & 64.31 & 40.27 & 75.30 & 138.00 \\
 & $\checkmark$ & \textcolor{gain}{86.73}
                & \textcolor{gain}{65.23}
                & \textcolor{gain}{40.91}
                & \textcolor{gain}{75.74}
                & \textcolor{gain}{140.86} \\
\bottomrule
\end{tabular}%
}
\end{minipage}
\hfill
\begin{minipage}[t]{0.49\textwidth}
\centering
\textbf{(b) Backbone size ablation}
\vspace{2pt}

\resizebox{\textwidth}{!}{%
\begin{tabular}{llccccc}
\toprule
Size & CoDeR & mIoU & BLEU-4 & MET. & R-L & CID. \\
\midrule
\multirow{2}{*}{Tiny}
 & $\times$     & 87.05 & 65.26 & 39.91 & 75.47 & 134.69 \\
 & $\checkmark$ & \textcolor{gain}{87.48} & \textcolor{gain}{65.76} & \textcolor{gain}{41.02} & \textcolor{gain}{76.13} & \textcolor{gain}{141.50} \\
\cmidrule(lr){2-7}
\multirow{2}{*}{Small}
 & $\times$     & 87.22 & 64.29 & 40.18 & 75.28 & 137.32 \\
 & $\checkmark$ & \textcolor{gain}{87.32} & \textcolor{gain}{65.30} & \textcolor{drop}{40.05} & \textcolor{gain}{75.33} & \textcolor{gain}{138.40} \\
\cmidrule(lr){2-7}
\multirow{2}{*}{Base}
 & $\times$     & 87.67 & 64.41 & 39.93 & 74.98 & 136.18 \\
 & $\checkmark$ & \textcolor{gain}{87.79} & \textbf{\textcolor{gain}{66.27}} & \textbf{\textcolor{gain}{41.08}} & \textbf{\textcolor{gain}{76.36}} & \textbf{\textcolor{gain}{141.79}} \\
\cmidrule(lr){2-7}
\multirow{2}{*}{Large}
 & $\times$     & 87.81 & 64.91 & 40.02 & 75.36 & 138.93 \\
 & $\checkmark$ & \textbf{\textcolor{gain}{87.85}} & \textcolor{gain}{65.73} & \textcolor{gain}{40.65} & \textcolor{gain}{76.01} & \textcolor{gain}{140.23} \\
\bottomrule
\end{tabular}%
}
\end{minipage}

\end{table*}

\begin{table*}[t]
\caption{Ablation studies on LEVIR-MCI~\cite{liu2024change}. 
(a) Single-scale versus multi-scale CoDeR application. The default design applies CoDeR at one receiving scale per direction (seg $\leftarrow$ cap at $1/8$, cap $\leftarrow$ seg at $1/16$), while the multi-scale variants apply independent CoDeR blocks at all four scales for the indicated direction(s). 
(b) Region-token pooling size ablation for caption decoding. \textbf{Bold} indicates the best result in each column.}
\label{tab:multiscale_region}
\centering
\small

\begin{minipage}[t]{0.54\textwidth}
\centering
\textbf{(a) Single-scale vs. multi-scale}
\vspace{2pt}

\resizebox{\textwidth}{!}{%
\begin{tabular}{llccccc}
\toprule
Type & Setting & mIoU & BLEU-4 & METEOR & ROUGE-L & CIDEr-D \\
\midrule
\multirow{3}{*}{Multi}
 & both     $(\text{all} \leftrightarrow \text{all})$      & 87.66 & 65.81 & 40.18 & 75.93 & 138.83 \\
 & seg only $(\text{all} \leftrightarrow 1/16)$            & 87.75 & 65.44 & 41.03 & 76.22 & \textbf{142.18} \\
 & cap only $(1/8 \leftrightarrow \text{all})$             & \textbf{87.86} & 64.56 & 40.83 & 75.93 & 140.32 \\
\midrule
Single (Ours) 
 & $(1/8 \leftrightarrow 1/16)$                            & 87.79 & \textbf{66.27} & \textbf{41.08} & \textbf{76.36} & 141.79 \\
\bottomrule
\end{tabular}%
}
\end{minipage}
\hfill
\begin{minipage}[t]{0.42\textwidth}
\centering
\textbf{(b) Region pooling size}
\vspace{2pt}

\resizebox{\textwidth}{!}{%
\begin{tabular}{lccccc}
\toprule
$M$ & mIoU & BLEU-4 & METEOR & ROUGE-L & CIDEr-D \\
\midrule
None & 87.56 & 65.95 & 40.80 & 75.99 & 139.22 \\
1    & 87.73 & 66.27 & 40.55 & 75.58 & 138.88 \\
2    & \textbf{87.79} & \textbf{66.27} & \textbf{41.08} & \textbf{76.36} & \textbf{141.79} \\
4    & 87.77 & 65.67 & \textbf{41.08} & 76.22 & 141.57 \\
\bottomrule
\end{tabular}%
}
\end{minipage}

\end{table*}

\begin{table*}[t]
\caption{Additional ablation studies on LEVIR-MCI~\cite{liu2024change}. \textbf{(a)} Effect of augmentation type, where $G$ denotes geometric augmentation and $R$ denotes directional caption remapping. \textbf{(b)} Effect of change fusion composition, where $F$ is formed from temporal features at $t_0$ and $t_1$ together with signed and magnitude change cues $\Delta$ and $|\Delta|$. \textbf{Bold} indicates the best result in each column.}
\label{tab:aug_comp}
\centering
\small

\begin{minipage}[t]{0.42\textwidth}
\centering
\textbf{(a) Augmentation type}
\vspace{2pt}

\resizebox{\textwidth}{!}{%
\begin{tabular}{lccccc}
\toprule
Aug. & mIoU & BLEU-4 & METEOR & ROUGE-L & CIDEr-D \\
\midrule
-   & 86.61 & 64.51 & 39.44 & 74.72 & 136.06 \\
G   & 86.67 & 64.45 & 40.64 & 75.97 & 139.21 \\
G+R & \textbf{87.79} & \textbf{66.27} & \textbf{41.08} & \textbf{76.36} & \textbf{141.79} \\
\bottomrule
\end{tabular}%
}
\end{minipage}
\hfill
\begin{minipage}[t]{0.55\textwidth}
\centering
\textbf{(b) Change fusion composition}
\vspace{2pt}

\resizebox{\textwidth}{!}{%
\begin{tabular}{lccccc}
\toprule
$F$ & mIoU & BLEU-4 & METEOR & ROUGE-L & CIDEr-D \\
\midrule
$\{t_0,t_1\}$                 & 87.45 & 64.21 & 40.04 & 75.37 & 138.76 \\
$\{t_0,t_1,\Delta\}$          & 87.47 & 65.69 & 40.53 & 75.87 & 140.86 \\
$\{t_0,t_1,\Delta,|\Delta|\}$ & \textbf{87.79} & \textbf{66.27} & \textbf{41.08} & \textbf{76.36} & \textbf{141.79} \\
\bottomrule
\end{tabular}%
}
\end{minipage}

\end{table*}

\begin{table*}[t]
\caption{Directional caption remapping under geometric augmentation on LEVIR-MCI. To preserve caption--image consistency, direction-sensitive expressions are remapped according to the applied geometric transform. Two-word directional phrases are matched prior to single-word directional levels.}
\label{tab:caption_remap_all}
\centering
\small
\resizebox{\textwidth}{!}{%
\begin{tabular}{lll}
\toprule
Transform & Category & Rule / Example \\
\midrule
\multicolumn{3}{c}{\textbf{Remapping rules}} \\
\midrule
\multirow{5}{*}{H-Flip}
& Single-level & $\textit{left} \leftrightarrow \textit{right}$ \\
& Phrase-level & $\textit{upper left} \leftrightarrow \textit{upper right}$ \\
& Phrase-level & $\textit{lower left} \leftrightarrow \textit{lower right}$ \\
& Phrase-level & $\textit{top left} \leftrightarrow \textit{top right}$ \\
& Phrase-level & $\textit{bottom left} \leftrightarrow \textit{bottom right}$ \\
\midrule
\multirow{4}{*}{V-Flip}
& Single-level & $\textit{top} \leftrightarrow \textit{bottom}$,\quad $\textit{upper} \leftrightarrow \textit{lower}$ \\
& Phrase-level & $\textit{top left} \leftrightarrow \textit{bottom left}$ \\
& Phrase-level & $\textit{top right} \leftrightarrow \textit{bottom right}$ \\
& Phrase-level & $\textit{upper left} \leftrightarrow \textit{lower left}$,\quad $\textit{upper right} \leftrightarrow \textit{lower right}$ \\
\midrule
\multirow{4}{*}{$90^{\circ}$ CCW rot. ($\times k$)}
& Single-level & $\textit{top} \rightarrow \textit{left} \rightarrow \textit{bottom} \rightarrow \textit{right} \rightarrow \textit{top}$ \\
& Single-level & $\textit{upper} \rightarrow \textit{left},\quad \textit{lower} \rightarrow \textit{right}$ \\
& Phrase-level & $\textit{top left} \rightarrow \textit{bottom left} \rightarrow \textit{bottom right} \rightarrow \textit{top right} \rightarrow \cdots$ \\
& Phrase-level & $\textit{upper left} \rightarrow \textit{lower left} \rightarrow \textit{lower right} \rightarrow \textit{upper right} \rightarrow \cdots$ \\
\midrule
\multicolumn{3}{c}{\textbf{Examples}} \\
\midrule
\multirow{6}{*}{Phrase-level}
& Original & \textit{a building appeared in the \textbf{upper left} corner of the scene} \\
& H-Flip & \textit{a building appeared in the \textbf{upper right} corner of the scene} \\
& V-Flip & \textit{a building appeared in the \textbf{lower left} corner of the scene} \\
& Rot.\ $90^{\circ}$ CCW & \textit{a building appeared in the \textbf{lower left} corner of the scene} \\
& Rot.\ $180^{\circ}$ & \textit{a building appeared in the \textbf{lower right} corner of the scene} \\
& Rot.\ $270^{\circ}$ CCW & \textit{a building appeared in the \textbf{upper right} corner of the scene} \\
\midrule
\multirow{3}{*}{Single-level}
& Original & \textit{the road on the \textbf{left} was removed} \\
& H-Flip & \textit{the road on the \textbf{right} was removed} \\
& Rot.\ $90^{\circ}$ CCW & \textit{the road on the \textbf{bottom} was removed} \\
\bottomrule
\end{tabular}%
}
\end{table*}

\subsection{Detailed Formulation of Feature Interaction Operators}
\label{sec:appendix_interaction_formulation}

This section provides the detailed formulations of the feature interaction operators compared in Table~\ref{tab:CoDeR_effective}(b), which serve as controlled baselines for cross-branch communication. The purpose of this comparison is not to faithfully reproduce the original architectures of prior methods, since they were introduced for different tasks, modalities, feature resolutions, and network designs. Instead, each operator is implemented as an abstraction of its core interaction principle within the same CoDeRNet framework. All variants are inserted at the same cross-task transfer locations and use the same encoder, task decoders, feature dimensions, and training protocol. This controlled setting allows the comparison to focus on the effect of the feature interaction design itself rather than on differences in the surrounding architecture.

Let $F_t, F_s \in \mathbb{R}^{B \times C \times H \times W}$ denote the target and source features, respectively. The target feature $F_t$ indicates the branch representation being updated, while the source feature $F_s$ provides complementary information from the other branch. For example, when the segmentation branch receives information from the captioning branch, $F_t$ corresponds to the segmentation feature and $F_s$ corresponds to the captioning feature. The reverse direction is defined analogously. We use $\phi(\cdot)$ to denote a lightweight source adapter implemented with a $1 \times 1$ convolution, and $\odot$ denotes element-wise multiplication. All operators are implemented as lightweight feature interaction blocks with comparable parameter counts and FLOPs.

\paragraph{Co-attention-style Interaction.}
The co-attention-style variant computes a spatial modulation map from the concatenated target and source features:
\[
a = \sigma\left(\mathrm{Conv}_{1\times1}([F_t;F_s])\right),
\]
where $[F_t;F_s]$ denotes channel-wise concatenation, $\sigma(\cdot)$ is the sigmoid function, and
$a \in [0,1]^{B \times 1 \times H \times W}$ is a spatial attention map. The target feature is updated by
\[
F_t' = F_t + a \odot \phi(F_s).
\]
This operator represents a co-attention-style spatial gating mechanism, where the source feature is injected into the target branch according to a spatial weight computed from the joint target-source representation. Since the target and source features are mixed before computing $a$, the resulting modulation does not explicitly distinguish receiver-side need for support from sender-side reliability.

\paragraph{Cross-attention Interaction.}
The cross-attention variant uses the target feature as query and the source feature as key and value. After flattening the spatial dimensions, we define
\[
Q = \mathrm{Flatten}(F_t), \quad K = \mathrm{Flatten}(F_s), \quad V = \mathrm{Flatten}(F_s),
\]
where $Q,K,V \in \mathbb{R}^{B \times HW \times C}$. The target feature is updated as
\[
F_t' = F_t + \mathrm{Reshape}\left(\mathrm{MHA}(Q,K,V)\right),
\]
where $\mathrm{MHA}(\cdot)$ denotes multi-head attention and $\mathrm{Reshape}(\cdot)$ restores the spatial feature shape. This operator captures token-wise target-source relations through attention weights. For a controlled comparison with the other lightweight operators, we use the attention operation as the interaction module without adding a full Transformer block with additional feed-forward layers.

\paragraph{Exchange-style Interaction.}
The exchange-style variant performs a soft channel-wise interpolation between the target feature and the adapted source feature. It first computes a channel-wise exchange gate:
\[
p = \sigma\left(\mathrm{Conv}_{1\times1}(\mathrm{GAP}(F_t))\right),
\]
where $\mathrm{GAP}(\cdot)$ denotes global average pooling and
$p \in [0,1]^{B \times C \times 1 \times 1}$. The target feature is updated by
\[
F_t' = (1-p) \odot F_t + p \odot \phi(F_s).
\]
This formulation can be interpreted as an exchange-inspired soft mixing operation: channels with small values of $p$ mainly preserve the target representation, while channels with large values of $p$ receive more information from the source branch. Unlike spatial gating, this operator controls feature transfer at the channel level and uses the same exchange weight across spatial locations. Unlike hard channel replacement strategies, this formulation implements a soft exchange-style interaction within the same comparison framework.

\paragraph{Gating-based Modulation.}
The gated modulation variant computes a channel-wise gate from the global context of both target and source features:
\[
z = \sigma\left(\mathrm{Conv}_{1\times1}(\mathrm{GAP}([F_t;F_s]))\right),
\]
where $z \in [0,1]^{B \times C \times 1 \times 1}$. The target feature is updated as
\[
F_t' = F_t + z \odot \phi(F_s).
\]
This operator performs residual source feature injection controlled by a channel-wise gate. Compared with exchange-style interaction, it preserves the original target feature through a residual update rather than interpolating between the target and source features.

\paragraph{Confidence-Decomposed Routing (CoDeR).}
CoDeR separately computes target-side sufficiency and source-side reliability cues:
\[
C_t = \mathcal{R}_t(F_t), \quad C_s = \mathcal{R}_s(F_s),
\]
where \(\mathcal{R}_t\) and \(\mathcal{R}_s\) are lightweight routing estimators implemented with \(1\times1\) convolution heads followed by sigmoid activation. The confidence-decomposed routing mask is defined as
\[
M = (1-C_t) \odot C_s.
\]
The target feature is then updated by
\[
F_t' = F_t + M \odot \phi(F_s).
\]
Here, $C_t$ and $C_s$ are learned routing cues rather than externally supervised confidence labels, estimating target-side sufficiency and source-side reliability, respectively. The routing mask \(M\) becomes large when the receiving branch requires support and the sending branch provides reliable evidence, encouraging source information to be transferred mainly where it is expected to be complementary. This differs from joint gating mechanisms that compute a single modulation weight from mixed target-source features, as CoDeR explicitly decomposes the transfer condition into receiver-side need and sender-side reliability terms.

\subsection{Change-Class IoU Evaluation}
\label{sec:ciou_evaluation}

To complement the multi-class mIoU evaluation on LEVIR-MCI, we additionally report change-class IoU (cIoU), which measures localization performance over the combined change regions. We merge the building and road classes into a single change class in both the predicted and ground-truth masks, and compute cIoU as the ratio of the total intersection to the total union across the test set. Unlike binary mIoU, which averages the IoUs of the background and change classes, cIoU evaluates only the change class. It therefore assesses change localization independently of the distinction between building and road changes.

Table~\ref{tab:ciou_statistical}(a) compares single-task training, joint training without CoDeR, and joint training with CoDeR. Joint training improves cIoU from 82.05 to 82.97 over the single-task baseline, and adding CoDeR further increases it to 83.07. The same trend holds for mIoU, with CoDeRNet achieving the highest scores on both metrics among the three settings. These results indicate improvements in both multi-class segmentation and localization of the merged change regions, although the incremental detection gains from CoDeR over joint training are modest: 0.12 percentage points in mIoU and 0.10 in cIoU.

\subsection{Statistical Analysis}
\label{sec:statistical_analysis}

To assess the statistical evidence for the detection improvements, we conduct an image-level paired bootstrap analysis comparing CoDeRNet with MCINet~\cite{liu2024change} on LEVIR-MCI and WHU-CDC. We use one fixed checkpoint per method and evaluate both methods on the same test images under the same evaluation protocol. In each bootstrap replicate, we sample image indices with replacement, drawing as many images as in the original test set, and apply the same sampled indices to both methods. We then pool the pixel-level confusion counts over the sampled images, recompute the dataset-level metrics, and calculate the performance differences between the two methods. We repeat this procedure 10,000 times and obtain 95\% confidence intervals using the 2.5th and 97.5th percentiles of the bootstrap distribution of performance differences.

Table~\ref{tab:ciou_statistical}(b) reports the performance differences computed on the full test sets and their bootstrap confidence intervals. Compared with MCINet, CoDeRNet improves mIoU and cIoU on LEVIR-MCI by 1.36 and 1.79 percentage points, respectively, and cIoU on WHU-CDC by 2.02 percentage points. The 95\% confidence intervals lie entirely above zero for all detection metrics reported in the table, supporting the observed improvements on both datasets under image-level resampling.

\begin{table*}[t]
\caption{Additional detection evaluation.
(a) Multi-class mIoU and change-class IoU (cIoU) on LEVIR-MCI~\cite{liu2024change}. Seg and Cap indicate change detection and change captioning training, respectively. cIoU is computed by merging building and road into a single change class. \textbf{Bold} indicates the best result in each metric column.
(b) Image-level paired bootstrap analysis. $\Delta$ denotes the full-test-set performance difference between CoDeRNet and MCINet. Differences and 95\% confidence intervals are expressed in percentage points. Positive values favor CoDeRNet.}
\label{tab:ciou_statistical}
\centering
\small

\begin{minipage}[t]{0.42\textwidth}
\centering
\textbf{(a) Change-class IoU evaluation}
\vspace{2pt}

\resizebox{\textwidth}{!}{%
\begin{tabular}{ccccc}
\toprule
Seg & Cap & CoDeR & mIoU & cIoU \\
\midrule
$\checkmark$ & $\times$ & $\times$ & 87.03 & 82.05 \\
$\checkmark$ & $\checkmark$ & $\times$ & 87.67 & 82.97 \\
$\checkmark$ & $\checkmark$ & $\checkmark$ & \textbf{87.79} & \textbf{83.07} \\
\bottomrule
\end{tabular}%
}
\end{minipage}
\hfill
\begin{minipage}[t]{0.54\textwidth}
\centering
\textbf{(b) Paired bootstrap analysis}
\vspace{2pt}

\resizebox{\textwidth}{!}{%
\begin{tabular}{llcc}
\toprule
Dataset & Metric & $\Delta$ & 95\% CI \\
\midrule
\multirow{2}{*}{LEVIR-MCI}
 & mIoU & $+1.36$ & $[+1.07,\,+1.70]$ \\
 & cIoU & $+1.79$ & $[+1.41,\,+2.17]$ \\
\midrule
WHU-CDC & cIoU & $+2.02$ & $[+0.57,\,+3.69]$ \\
\bottomrule
\end{tabular}%
}
\end{minipage}

\end{table*}

\begin{figure}[t]
    \centering
    \includegraphics[width=\linewidth]{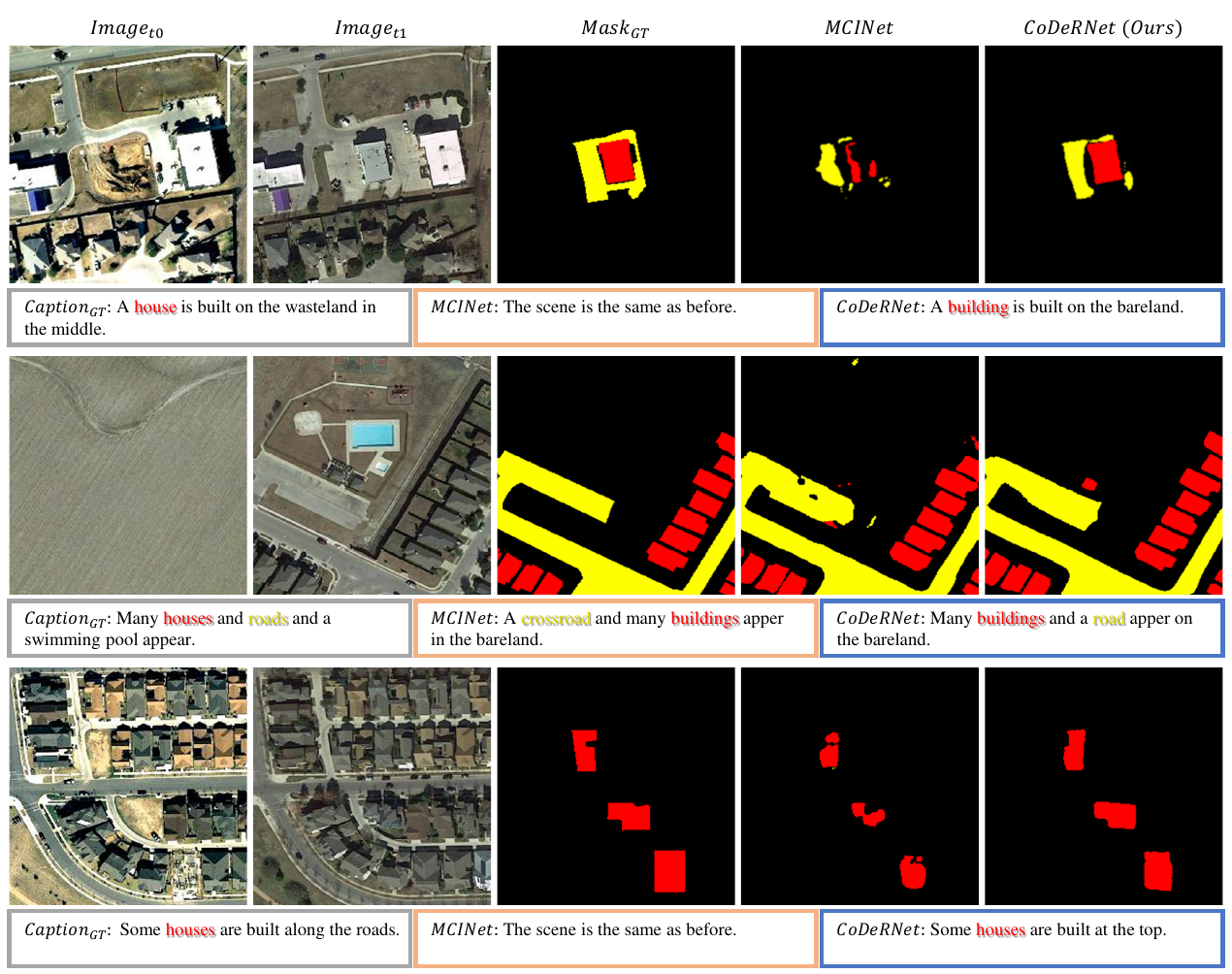}
\caption{Qualitative comparison on LEVIR-MCI. In the captions, building-related words (e.g., \textit{house}, \textit{building}) and road-related words (e.g., \textit{road}) are highlighted using colors consistent with the corresponding mask classes.}
    \label{fig:qualitative_levir2}
\end{figure}
\begin{figure}[t]
    \centering
    \includegraphics[width=\linewidth]{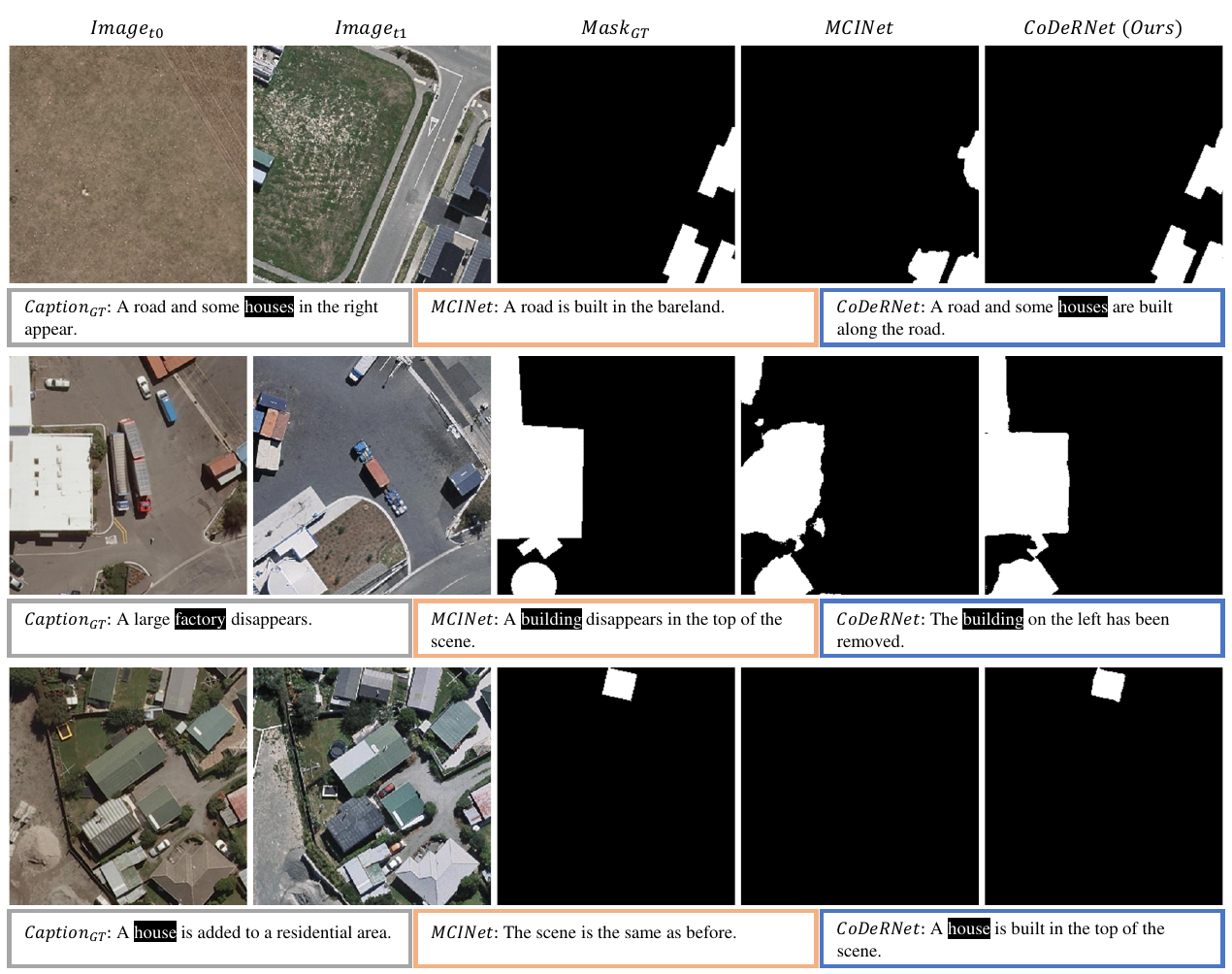}
\caption{Qualitative comparison on WHU-CDC. In the captions, building-related words (e.g., \textit{house}, \textit{building}) are highlighted using colors consistent with the corresponding mask classes.}
    \label{fig:qualitative_whu}
\end{figure}

\newpage

\clearpage

\end{document}